\pdfoutput=1

\documentclass[11pt]{article}

\usepackage[final]{emnlp2021}

\usepackage{times}
\usepackage{latexsym}
\usepackage[T1]{fontenc}

\usepackage[utf8]{inputenc}

\usepackage{booktabs}

\usepackage{microtype}

\newcommand{\BB}[0]{BERT\textsubscript{BASE}}
\newcommand{\BL}[0]{BERT\textsubscript{LARGE}}
\newcommand{\RB}[0]{RoBERTa\textsubscript{BASE}}
\newcommand{\RL}[0]{RoBERTa\textsubscript{LARGE}}
\newcommand{\AB}[0]{ALBERT\textsubscript{BASE}}
\newcommand{\AL}[0]{ALBERT\textsubscript{LARGE}}
\newcommand{\AXL}[0]{ALBERT\textsubscript{XLARGE}}
\newcommand{\AXXL}[0]{ALBERT\textsubscript{XXLARGE}}

\title{Pragmatic competence of pre-trained language models through the lens of discourse connectives}

  \author{Lalchand Pandia, Yan Cong$^1$, Allyson Ettinger$^2$ \\
  $^1$Department of Linguistics and Languages, Michigan State University \\
  $^2$Department of Linguistics, University of Chicago \\
  \texttt{lcpandia@gmail.com}, \texttt{congyan@msu.edu}, \texttt{aettinger@uchicago.edu} \\ 
  }

\begin{document}
\maketitle 
\begin{abstract}
As pre-trained language models (LMs) continue to dominate NLP, it is increasingly important that we understand the depth of language capabilities in these models. In this paper, we target pre-trained LMs' competence in pragmatics, with a focus on pragmatics relating to discourse connectives. We formulate cloze-style tests using a combination of naturally-occurring data and controlled inputs drawn from psycholinguistics. We focus on testing models' ability to use pragmatic cues to predict discourse connectives, models' ability to understand implicatures relating to connectives, and the extent to which models show humanlike preferences regarding temporal dynamics of connectives. We find that although models predict connectives reasonably well in the context of naturally-occurring data, when we control contexts to isolate high-level pragmatic cues, model sensitivity is much lower. Models also do not show substantial humanlike temporal preferences. Overall, the findings suggest that at present, dominant pre-training paradigms do not result in substantial pragmatic competence in our models.
\end{abstract}

\section{Introduction}

Pre-trained language models continue to display impressive performance across various domains of NLP, raising the important question of exactly what level of linguistic competence these models have acquired, particularly during the pre-training process. Although models show outstanding performance on downstream tasks, there is also evidence that their handling of language is shallow~\cite{mccoyEtal2019,sinha2021masked}. 

In this paper we examine aspects of pre-trained LMs competence in pragmatics, with a particular focus on pragmatic reasoning surrounding discourse connectives. Discourse connectives are linguistic elements that connect two neighboring events or sentences, signalling the discourse relation that exists between them. 
Discourse connectives reflect and convey pragmatic information about relationships between events being described, and connectives can also be associated with pragmatically-enriched meanings that go beyond their literal meanings. If pre-trained LMs have acquired competence in pragmatics of a language, we would expect them to be able to use cues in context to infer what discourse relation holds between events, and by extension which discourse connective is appropriate. Additionally, we would expect them to be able to pick up on pragmatic inferences generated by the connectives themselves. These are the questions that we test in this paper. 


We formulate all of our tests as cloze tasks, so that we can test the pre-trained LMs without fine-tuning. We begin with a general task of connective prediction, to gauge how well models can use surrounding sentence context to predict the appropriate discourse connective. 
We then move on to more controlled tests, inspired by psycholinguistics, to ask more targeted questions about models' use of contextual cues and handling of implicature. The first of these examines models' ability to use high-level pragmatic cues to infer appropriate connectives, in the absence of clear syntactic or lexical cues. The second examines the extent to which model predictions reflect understanding of temporal implicatures in settings of implicature cancellation and reinforcement. Finally, we test whether models show humanlike preferences in terms of  temporal dynamics of event mentions. 

We test a range of pre-trained LMs with these analyses. Our results indicate that although models show reasonable connective predictions in naturally-occurring data, they lack the pragmatic sensitivity to perform well on our controlled tests, and they don't show any strong humanlike preference in terms of temporal dynamics. The results suggest that for the moment, the currently dominant pre-training paradigms are not yielding clear pragmatic competence in NLP models, at least with respect to discourse connectives. We make all code and test data available for additional testing.\footnote{\url{https://github.com/lalchand-pandia/Pragmatic-competence-in-PLM}}



\section{Related Work}


Connectives have been studied in NLP models from a number of angles. Some works have focused on improving automatic extraction of connectives via heuristics~\cite{sileo-etal-2019-mining1} or dependency parses~\cite{nie-etal-2019-dissent1}, so as to use connective prediction during model pre-training. While these works share with ours the use of discourse connective prediction, we differ in focusing not on modifying model training, but on evaluating and analyzing existing models, to better understand the extent to which existing pre-training paradigms confer sensitivity to pragmatic cues.
 

Others have studied models' competence in classifying discourse relations on the basis of connectives or contexts. ~\citet{Pitler2009UsingST} show that using syntactic features along with the connectives themselves, a supervised classifier is able to identify the discourse relation that a connective represents.
~\citet{kurfali-ostling-2021-discourse} study competence of multilingual models in various discourse tasks by evaluating cross-lingual zero-shot transfer in a range of sentence encoders---among these tasks is classification of discourse relations, though they focus on implicit discourse relations, where connectives are not the source of discourse information. ~\citet{koto-etal-2021-discourse} also evaluate pre-trained language models' performance on discourse level relations, including a task of discourse connective prediction: given two sentences, the task is to predict the explicit connective. They fine-tune pre-trained models on this classification task,
finding that for the discourse connective prediction task, different models produce similar patterns. 
\citet{Patterson2013PredictingTP} also examine connective prediction, training a classifier to predict whether a connective is present, and finding that classifiers are able to perform this prediction task on the basis of shallow linguistic features alone.

Our work in this paper is similar to these prior works in examining the ability of models to capture discourse information through tasks involving prediction of connectives and discourse relations. However, we differ critically from prior work in asking not whether we can train models to make these predictions---rather, our question is to what extent pre-trained LMs have already developed pragmatic knowledge relating to connectives, as a byproduct of the pre-training process itself. Consequently, we deviate from this prior work in that we do not fine-tune models or do any supervised training for connective predictions---instead, we formulate our tests as word prediction tasks and test whether model predictions reflect an understanding of the pragmatics surrounding connectives. 

Our work also distinguishes itself from prior work on connectives in that we anchor our tests in insights and methods from neuro- and psycho-linguistic experiments, which are well controlled and designed for assessing linguistic behavior and competence in a targeted manner. This helps us to tease out potential superficial cues that may inflate perceived levels of pragmatic competence. Modifications that we make to the original psycholinguistic experimental items are furthermore grounded in established findings of pragmatic theory.

\section{Experiments}

\subsection{Models}

We apply our tests to examine three classes of pre-trained LMs, testing various size settings within each class. For the models analyzed in this paper, we use the implementation of~\citet{wolf2020huggingfaces}. We limit our investigation to masked language models, since it is necessary that models be able to use right-hand context for predicting connectives.

\paragraph{BERT~\cite{devlin2019bert}} We experiment with two variants: \BB{} (110M parameters), and \BL{} (340M parameters). For both, we use the uncased version.

\paragraph{RoBERTa~\cite{liu2019roberta}} We experiment with \RB{} (125M parameters) and \RL{} (355M parameters).
\paragraph{ALBERT~\cite{Lan2020ALBERTAL}} We experiment with version 2 of \AB{} (11M parameters), \AL{} (17M parameters, \AXL{} (58M parameters) and \AXXL{} (223M parameters).

\subsection{Input representation}

For our inputs, we add a start of sentence token (\emph{$[$CLS$]$} for BERT, ALBERT; \emph{<s>} for RoBERTa). Separate sentences of a given input item are separated by a separator token, and the masked word to be predicted is denoted by \emph{$[$MASK$]$} for BERT and ALBERT, and \emph{<mask>} for RoBERTa. Special tokens are selected for consistency with the implementation of~\citet{wolf2020huggingfaces}.\footnote{One reviewer raised a concern about inputs of more than two clauses/sentences, as in Section~\ref{sec:drenhaus}. Note that such multi-sentence inputs are consistent with models' pre-training, during which input ``sentences'' are not defined by actual sentence boundaries, but by selection of arbitrary spans of contiguous text, thus allowing for multi-sentence inputs.}

\begin{table}[t!]
\centering
\begin{tabular}{l|l}
\toprule
  Model & Accuracy\\
  \midrule
  \BB &0.47\\
  \BL & 0.51\\
  \RB & 0.61\\
  \RL & 0.66\\
  \AB & 0.42\\
  \AL & 0.48\\
  \AXL & 0.56\\
  \AXXL & 0.57\\
  \bottomrule
\end{tabular}
\caption{Connective prediction accuracy on PDTB data}
\label{table:Acc_PDTB}
\end{table}

\begin{table*}[ht!]
\centering
\begin{tabular}{l|l l l l}
\toprule
  Model & Expansion:&Asynchronous: & Concession: & Causal:\\
  & Conjunction&Succession&contra-expectation&Result\\
  \midrule
  \BB &0.73& 0.46 & 0.18&0.20\\
  \BL & 0.74&0.5&0.25&0.24\\
  \RB & 0.76&0.67&0.43&0.3\\
  \RL & 0.79&0.71&0.51&0.34\\
  \AB & 0.42& 0.52& 0.37&0.09\\
  \AL & 0.49& 0.61& 0.38& 0.17\\
  \AXL &0.64 & 0.62& 0.41&0.25\\
  \AXXL &0.59 & 0.66& 0.46&0.29\\
  \bottomrule
\end{tabular}
\caption{Connective prediction accuracy on PDTB data, broken down by specific discourse relations}
\label{table:Acc_PDTB_relation_wise}
\end{table*}

\section{Predicting connectives in naturally-occurring data}\label{sec:pdtb}

We begin by asking, generally speaking, how effective pre-trained LMs are at using context to infer the appropriate connective to join components of a given discourse. With this experiment, we take advantage of large amounts of naturally-occurring data to gauge the general capacities of these models to use surrounding information to infer the most appropriate discourse connective.

We compile 17,476 input items, drawn from instances in the Penn Discourse Treebank (PDTB-2)~\cite{prasad-etal-2008-penn1}. We select instances based on presence of explicit discourse connectives, according to dataset annotations. We filter out any instances in which connectives are multi-word, to enable use of a masked single-word prediction setting. For testing, we use a cloze approach, simply masking the discourse connective and assessing the probabilities that the masked language models assign to the correct connective given the context. The models receive only a single PDTB instance at a time as input. We measure prediction accuracy in relative terms: models are considered accurate if they assign a higher probability to the correct connective than to any other single-word connectives in PDTB (66 candidate connectives in total).


We note that of course even humans may struggle to predict many naturally-occurring connectives, and a given context may be consistent with multiple discourse relations.\footnote{It has also been observed PDTB-2 includes connectives that can signal more than one discourse relation~\citep{Pitler2009UsingST, webber-etal-2019-ambiguity1}. PDTB-3 tries to resolve this connective ambiguity by introducing new relations in the annotations, but for the purposes of our preliminary test here, PDTB-2 is sufficient.} 
This relative lack of control over item properties is a tradeoff that comes with use of large-scale naturally-occurring data, but the level of predictability in these items can be assumed to mirror levels of predictability in the models' normal pre-training. More to the point, however, the experiments in this section serve only as a preliminary assessment of models' ability to use contextual information to infer discourse relations and corresponding connectives used in original texts. Subsequent sections will shift to more targeted tests of how models handle the pragmatics of connectives. Note also that the difficulty of predicting connectives here would if anything lead accuracies in this section to be underestimates---but even with this disadvantage, we will see that these PDTB accuracies still appear to overestimate models' actual pragmatic competence. This further highlights the need for more controlled tests.


Table~\ref{table:Acc_PDTB} shows the overall accuracy results for these PDTB items. 
We see that models prefer the correct connective to other connectives approximately half the time---well above chance---suggesting reasonable ability to use contextual cues to infer the appropriate discourse connective. Accuracy generally improves with model size within model class, and among the three model classes, RoBERTa shows the strongest performance overall. If we take these results at face value, it appears that models may have a reasonable grasp on pragmatics of connectives---and increasing a given model size may improve pragmatic competence still further. 


When we break accuracies down by relation types, however, we find that accuracies vary drastically among different relations. In Table~\ref{table:Acc_PDTB_relation_wise} we show the accuracy of selected relation types of Expansion.Conjunction (which can be signalled by connectives like \textit{and, also, additionally}), Comparison.Concession.Contra-expectation (signalled by connectives like \textit{but, however, although}), Temporal.Asynchronous.Succession (signalled by connectives like \textit{after, since, when}) and Causal (signalled by \textit{so, thus, therefore}). Comparing between relation types, we see that models (particularly BERT and RoBERTa) show much higher accuracy on Expansion.Conjunction,
more moderate performance on Asynchronous.Succession and Concession, and generally quite weak performance at predicting Causal connectives. ALBERT deviates somewhat from the pattern of BERT and RoBERTa, in that its highest performance is instead typically on Asynchronous.Succession.




These slightly lopsided accuracies suggest a picture in which LMs may make some use of pragmatic contextual cues, but they may also have certain common, go-to connectives that serve as probable predictions across a wide range of contexts. Error analysis is consistent with this picture---Tables ~\ref{table:error_percentage_of_connectives_PDTB_bert} and ~\ref{table:error_percentage_of_connectives_PDTB_albert} in the appendix show percentages of erroneous predictions for which each candidate connective is the top-ranked prediction. We see that across the board, BERT and RoBERTa models have high rates of preferring \emph{and} in cases of erroneous prediction. ALBERT, by contrast, distributes errors across a wider range of connectives. What we see then, is a picture in which BERT and RoBERTa models seem to have settled on \emph{and} as a common go-to connective, contributing to the high accuracy of those models on the Expansion.Conjunction relation---while ALBERT has less of a go-to \emph{and} preference, consistent with ALBERT's lower accuracy on Expansion.Conjunction, and slightly more balanced accuracies overall. 

What do these results mean for our assessment of these LMs? The strategic benefit of frequently predicting \emph{and} is clear: \emph{and} is a versatile, ambiguous discourse connective that can appear in many types of contexts, so it is probably among the safest connective predictions. However, this rather coarse-grained predictive behavior suggests that models may not be very sensitive to detailed pragmatic cues that would enable more specific connective predictions that fit the contexts more precisely. Since we have not controlled the contexts on which the models condition here, we cannot make strong claims about the specific cues that models may or may not have had access to for each of these individual relations. To address this, below we will use controlled sentence contexts aimed at isolating pragmatic information for a more targeted test of connective prediction capabilities. The first of these tests will focus on distinguishing causal and concessive connective environments.

These PDTB tests are also limited in what they can tell us about models' understanding of the \emph{meanings} of connectives---in particular, given models' inclination to over-predict \emph{and}, it is difficult to know the extent to which models have any understanding of the implications of this connective in context. We will look further into this question below, by isolating a particular temporal implicature of the connective \emph{and}, and testing whether models can predict other temporal connectives reflecting the meaning of that implicature. This test will make use of the influences of two hallmarks of implicature---reinforcement and cancellation---to test models' pragmatic sensitivities. Finally, we will further probe models' sensitivity to temporal dynamics of connectives, by testing whether models' connective predictions reflect a humanlike preference for events to be mentioned in the order in which they occur in the real world.


For all of these follow-up experiments, we will make use of insights and tests from psycholinguistics---this will enable us to execute more controlled and targeted tests, while grounding our expectations in observed properties of human processing and interpretation of connectives. Adaptations of the original experimental items will be grounded in insights from pragmatic theory.


\begin{table*}[ht]
\centering
\begin{tabular}{p{.2\textwidth}|p{.7\textwidth}}
\toprule
Condition & Example item \\
\midrule
Causal connective context & \emph{Mr. Brown was planning to look for new glasses and shoes today. The glasses really are more urgent. \textbf{\underline{[MASK]}}, he now heads towards the \textbf{\underline{optician}} that a friend recommended.} \textbf{(correct target: \emph{therefore, so})} \\ \midrule
Concessive connective context & \emph{Mr. Brown was planning to look for new glasses and shoes today. The glasses really are more urgent. \textbf{\underline{[MASK]}}, he now heads towards the \textbf{\underline{shoe store}} that a friend recommended.} \textbf{(correct target: \emph{however, but})} \\
\bottomrule
\end{tabular}
\caption{Example items from~\citet{Drenhaus-etal-2014}, adapted for testing connective prediction in controlled contexts. Models should be able to use pragmatic cues to infer appropriateness of causal vs concessive connective.}\label{tab:drenhausitems}
\end{table*}

\begin{table}[t!]
\centering
\begin{tabular}{l|l l l}
\toprule
  Model &  Conces. & Causal & Pair\\
  \midrule
  \BB &  0.73&0.27&0\\
   \BL & 0.6&0.43&0.03\\
   \RB & 0.4&0.46&0\\
   \RL & 0.43&0.67&0.1\\
    \AB &0.9& 0.1&0\\
   \AL & 0.6&0.4&0.03\\
   \AXL & 0.53 &0.5&0.07\\
   \AXXL &  0.57&0.7&0.3\\
  \bottomrule
\end{tabular}
\caption{Prediction accuracy on contexts from \citet{Drenhaus-etal-2014}. ``Conces'' = Concessive; ``Pair'' = rate of correct prediction on both sentences of a pair}
\label{table:Acc_Drenhaus}
\end{table}

\section{Predicting connectives with controlled context} \label{sec:drenhaus}

The results in Section~\ref{sec:pdtb} give a preliminary sense of models' behavior in using context to predict connectives, but it is difficult to discern from these uncontrolled data precisely what types of cues the models may be using to inform connective predictions. 
In particular, if we are interested in models' ability to use high-level pragmatic information to infer the appropriate discourse relation, it is important that we control for lower-level syntactic and lexical cues that may be predictive of connectives, but that tell us less about models' sensitivity to pragmatics. Previous works have shown that lexical and semantic cues can be used for predicting connectives and discourse relations~\cite{Patterson2013PredictingTP,Pitler2009UsingST}, and that certain kinds of relations co-occur at rates greater than chance~\cite{Pitler2008EasilyID}, supporting the possibility that non-pragmatic cues alone can likely lead to strong connective prediction performance. 



In order to better isolate high-level pragmatic cues, we take advantage of sentences designed by~\citet{Drenhaus-etal-2014} for a psycholinguistic study of human language processing. The original psycholinguistic experiment tested how different discourse connectives facilitate human language comprehension, and the extent to which connectives can elicit predictions of upcoming content.
The experimental items constitute minimal pairs with nearly identical syntax and word content---but a slight difference late in the context makes it such that a causal connective is appropriate in one version, while a concessive connective is appropriate in the other.\footnote{Concessive relation is equivalent to PDTB COMPARISON:Concession:contra-expectation; causal relation is equivalent to CONTINGENCY:Cause:result.} Taking advantage of this controlled minimal pair set-up, we adapt the items from this study to formulate a connective prediction task---Table~\ref{tab:drenhausitems} shows examples from these adapted items. To do this task, the models need to identify that in one context Mr. Brown's actions follow what is expected from the first sentence, while in the other context, the actions deviate from what is expected. Our goal is to test whether models can use pragmatic reasoning to infer that these contexts are conducive to a causal and a concessive connective, respectively.  In filtering these items, we again leave out connectives that are multi-word. We derive a total of 30 item pairs for these experiments.\footnote{In each item, the sentence containing the mask token is separated from the preceding sentence with a \textit{[SEP]} token.}


When testing on these items, we consider a model to be accurate if, from a list of causal and concessive connectives, the model's top prediction falls in the correct category (causal versus concessive). For causal connectives we count any of \{\textit{so, therefore}\}, and for concessive connectives we count any of \{\textit{however, instead, nevertheless}\}. 

Table~\ref{table:Acc_Drenhaus} shows model prediction accuracy on this test. When we look at accuracy for concessive and causal sentences separately, we see that certain models do appear to have strong performance. However, performance on a single condition is again susceptible to models simply preferring certain connectives in both contexts, and we are interested in models' ability to use subtle pragmatic information to distinguish the minimal pairs. Thus, we focus on the proportion of item pairs on which the models manage to prefer the correct class of connective in both items of the pair. This is shown in the ``Pair'' column of Table~\ref{table:Acc_Drenhaus}, and it is clear that models perform extremely poorly by this criterion. Most models hover around 0\% accuracy---only \AXXL{} exceeds (narrowly) the roughly 25\% threshold that would be expected by chance. The driving reason for these failures is the fact that for a given minimal pair, models continue to prefer the same completion in both items, failing to respond to the subtle changes in contextual pragmatic cues. On the whole, the results suggest that when we limit models' cues to high-level pragmatic information of the type targeted in these items, none of these tested models have the capacity to use that information to distinguish and predict both causal and concessive connectives. Models' difficulty on this test can also be seen as somewhat consistent with the findings from Section~\ref{sec:pdtb} that models are weaker in general on causal and concessive connectives---however, the much more dramatic failure here may indicate that where we do see correct predictions in Section~\ref{sec:pdtb}, those predictions may be informed by shallower cues, rather than subtle pragmatic information.


\begin{table*}[ht]
\centering
\begin{tabular}{p{.2\textwidth}|p{.7\textwidth}}
\toprule
Condition & Example item \\
\midrule
\bottomrule
Reinforcement test & \emph{Maggie did the paperwork by hand \textbf{\underline{and}} the company bought new computers, which is to say, Maggie did the paperwork by hand \textbf{\underline{[MASK]}} the company bought new computers.} \textbf{(ideal target probs: \emph{before} $>$ \emph{after})}\\ \midrule
Cancellation test & \emph{Maggie did the paperwork by hand \textbf{\underline{and}} the company bought new computers, in fact, Maggie did the paperwork by hand \textbf{\underline{[MASK]}} the company bought new computers.} \textbf{(ideal target probs: \emph{after} $>$ \emph{before})}\\
\bottomrule
\end{tabular}
\caption{Example items for testing interpretation of \emph{and then} implicature, using reinforcement and cancellation}\label{tab:reincancelitems}
\end{table*}

\begin{table}[t!]
\centering
\begin{tabular}{l|l l l }
\toprule
  Model & Cancel. & Reinf. & Pair \\
  \midrule
  \BB & 0.52 & 0.55&0.08\\
  \BL & 0.24&0.79&0.04\\
  \RB & 0.69&0.49&0.18\\
  \RL & 0.61&0.21&0.04\\
  \AB &0.66 &0.3 & 0.02\\
  \AL & 0.83& 0.37&0.2 \\
  \AXL & 0.06& 0.98& 0.04\\
  \AXXL &0.09 & 0.89& 0.06\\
  \bottomrule
\end{tabular}
\caption{Prediction accuracy in implicature test, with cancellation and reinforcement settings}
\label{table:Acc_Modified_Drenhaus}
\end{table}

\section{Discourse connectives and implicature}

Section~\ref{sec:pdtb} suggests that models are quick to predict \emph{and} in contexts that generally support a discourse connective. Here we test the extent to which models understand what the connective \emph{and} actually means---more specifically, we examine the extent to which models pick up on temporal implicatures that humans commonly interpret as part of the meaning of \emph{and}. This section will again use controlled minimal pairs of contexts, inspired by findings in linguistics and psycholinguistics.

Our tests here make use of the pragmatic notion of \emph{implicature}: non-literal, enriched meaning of words, phrases, or sentences generated by inferences about speaker intent.  For instance, the sentence ``Some people have pets'' literally means that ``there are a non-zero number of people who have pets''. In addition to this literal meaning, a common implicature in interpretation of this sentence is that ``Some but not all people have pets''.



We focus on an implicature generated by the connective \emph{and} when joining two events---namely, the implicature that \emph{and} actually means \emph{and then} (i.e., the two events are being mentioned in the temporal order in which they actually occurred). This has been studied by~\citet{Carston1988-CARIEA-7}, noting the oddness of sentences like ``Jane got into bed and brushed her teeth'', which seems to carry the clear implication that Jane brushed her teeth in bed. ~\citet{NoveckChevaux2002} also study this implicature in children, finding that compared to younger children, older children and adults generate more \emph{and then} implicatures when events are joined by \emph{and}. Our tests will focus first on whether models seem to be sensitive to this \emph{and then} implicature.

To test whether models pick up on this implicature, we cannot simply test models' aptitude at predicting the connective \emph{and} in context. Instead, we make use of two additional hallmarks of implicatures~\cite{Grice1975,grice1989}: 1) that they can be reinforced with, e.g., ``which is to say'', and 2) that they can be canceled with, e.g., ``in fact''. (For instance, we can reinforce the ``some but not all'' implicature above by saying, ``Some people have pets---which is to say, some, \textit{but not all}, people have pets''. Alternatively, we can cancel the ``some but not all'' implicature by saying ``Some people have pets---in fact, \textit{all} people have pets''.) These tests are well established in the linguistics literature for teasing apart implied meanings from compositional truth conditions~\citep{fox2007, katzir2007, geurts2010,chierchia.etal2012, sauerland2004,sadock1978,rett2014semantics}. We will leverage sentences featuring reinforcement and cancellation of the \emph{and then} implicature to test our models.


We create a dataset of item pairs containing events joined by \emph{and}, followed by a reinforcement or cancellation. Table~\ref{tab:reincancelitems} shows example items. We draw our events from the stimuli of~\citet{PolitzerAhles2017BeforeA}, which tested humans' sensitivity to temporal order of events (in this section we simply insert events from those stimuli into sentences of our chosen structure---in Section~\ref{sec:politzerahles} we will make more direct use of the stimuli from that study). 
We create 160 item pairs in total for use in this test.


When we examine model predictions in the masked positions of these items, the critical question is the relative probability that they assign to completions of \emph{before} versus \emph{after}. If models generate the natural \emph{and then} implicature for the connective \emph{and} (and if they have the pragmatic competence to understand the effects of ``which is to say'' and ``in fact''), then they should prefer \emph{before} in the case of reinforcement, and \emph{after} in the case of cancellation. This approach to assessment is particularly important because these contexts are somewhat complex, and the human standard to which we are comparing models in this case is not direct human performance on these items, but rather related results and theoretical foundation in the pragmatics literature. For these reasons we seek to maximize the fairness of the test through use of this relative accuracy: to be considered correct, models need only prefer the temporal relation that better fits the invoked implicature, over the temporal relation that clashes with the invoked implicature.

\begin{table*}[ht]
\centering
\begin{tabular}{p{.2\textwidth}|p{.7\textwidth}}
\toprule
Condition & Example item \\
\midrule
Sentence-initial & \emph{\textbf{\underline{[MASK]}} the campaign finance laws changed, Albert ran for mayor of his city.} \textbf{(temporal order preference: \emph{after} $>$ \emph{before})}\\ \midrule
Sentence-medial & \emph{Albert ran for mayor of his city \textbf{\underline{[MASK]}} the campaign finance laws changed.} \textbf{(temporal order preference: \emph{before} $>$ \emph{after})}\\
\bottomrule
\end{tabular}
\caption{Example items from \citet{PolitzerAhles2017BeforeA}, adapted for testing whether models prefer connectives that indicate events are being mentioned in their chronological order. When connective is at the beginning of the sentence, \emph{after} indicates that events will be mentioned in chronological order. When the connective is in the middle of the sentence, chronological ordering is signalled by \emph{before}.}\label{tab:politzeritems}
\end{table*}

\begin{table}[t!]
\centering
\begin{tabular}{l|lll}
\toprule
  Model & Initial & Medial & Pair\\
  \midrule
  \BB & 0.93 & 0.45&0.38\\
  \BL & 0.91&0.48&0.39\\
  \RB & 0.85&0.39&0.24\\
  \RL & 0.86&0.4&0.26\\
   \AB & 0.95& 0.46& 0.41\\
  \AL & 0.95& 0.24&0.2\\
  \AXL & 0.75&0.54 &0.31\\
  \AXXL & 0.68& 0.56&0.26\\
  \bottomrule
\end{tabular}
\caption{Model preferences for \emph{after}/\emph{before} in sentence-initial and sentence-medial position. ``Initial'' shows percentage of sentence-initial predictions that prefer \emph{after}, while ``Medial'' shows percentage of sentence-medial predictions that prefer \emph{before}.}
\label{table:Acc_Politzer_Before_After}
\end{table}



Table~\ref{table:Acc_Modified_Drenhaus} shows the results. The ``Cancel.'' column shows the proportion of \textit{in fact} items in which the model assigns higher probability to \textit{after} than to \textit{before}, and the ``Reinf.'' column shows the proportion of \textit{which is to say} items in which the model assigns higher probability to \textit{before} than to \textit{after}. As before, these single-condition accuracies are susceptible to models simply preferring one connective across contexts, so we are most interested in the ``Pair'' column, which shows the proportion of minimal pairs (reinforcement + cancellation version) in which the model assigns higher probability to the correct target for both items. 

As in Section~\ref{sec:drenhaus}, we see that although models may make correct predictions on individual items, their ability to choose the correct connective in both items of a pair is very limited. Chance-level performance on this criterion is 25\%, and it is clear that most models are performing substantially below chance level---and even the highest accuracies remain slightly below chance. The results indicate that this form of pragmatic competence---inferring temporal implicature and deploying it in cancellation and reinforcement environments---remains outside of models' current capacity.


\subsection{Sensitivity to redundancy and contradiction}

As a follow-up to the implicature test above, we also test how model behaviors change when the implied meaning is made explicit. We pair each item from the prior experiment (Table~\ref{tab:reincancelitems}) with a counterpart containing \emph{and then} in place of \emph{and}. This time we focus on a single candidate prediction word at once, and compare the probability of that word in \emph{and} versus \emph{and then} versions of a given item. In cancellation conditions, we examine how models rate a target of \emph{after}---models should assign higher probability to \emph{after} when the sentence contains only \emph{and}, because \emph{and then} is contradictory with an \emph{after} interpretation. In reinforcement sentences, we examine model probabilities for \emph{before}---models should assign higher probability to \emph{before} in reinforcement sentences with only \emph{and}, because restating that an event X happened before an event Y is redundant if \emph{and then} is stated explicitly. Example items from this test are listed in appendix Table ~\ref{tab:reincancelitemssup}. We create 320 sentence pairs in total for this test.\footnote{The number of items is doubled relative to the previous experiment because we now have both an \emph{and then} version and an \emph{and} version of each of the original items.} We count model predictions as correct if the target word of interest is assigned higher probability in the appropriate context than in the inappropriate context, as outlined above.







For the sake of space, we show the results in Table~\ref{table:Acc_Modified_Drenhaus_Cancellability} of the appendix. Model performance is extremely weak across the board, with models assigning higher probability in the better context no more than 3\% of the time---except for \RL{} in Reinforcement conditions, at 21\% (chance level of 50\%). Overall, the results suggest that models are sensitive neither to contradiction nor to redundancy---or at least that they prefer sentences with these properties over sentences featuring reinforcement and cancellation of an implicature.




\section{Sensitivity to event order and corresponding connectives} \label{sec:politzerahles}

The previous section tested whether models, like humans, infer that a connective \emph{and} joining two events has an implied meaning of \emph{and then}. A related finding in psycholinguistics is from~\citet{PolitzerAhles2017BeforeA}, who examine the effect of \textit{before} and \textit{after} clauses on sentence processing in both sentence-initial contexts and sentence-medial contexts. Their results provide further evidence that human brains have a preference for events to be mentioned in chronological order. We leverage this existing experiment to probe whether models, in their connective predictions, show similar preferences for events to be mentioned in the order in which they occur in real life. Importantly, this test differs from the above tests in that models cannot be said to behave ``correctly'' or ``incorrectly'', on the basis of the presence/absence of this preference. Rather, this serves as a more general test of whether models' connective predictions reflect humanlike trends with respect to temporal implications.

We adapt the~\citet{PolitzerAhles2017BeforeA} materials to form a connective prediction task. Examples are shown in Table~\ref{tab:politzeritems}. The central question is whether models will prefer connectives that imply that events are being mentioned in chronological order. Models are considered to have this preference if they assign higher probability to \emph{after} (relative to \emph{before}) in sentence-initial position, and to \emph{before} (relative to \emph{after}) in sentence-medial position. 




Table~\ref{table:Acc_Politzer_Before_After} shows the results. The patterns suggest that models strongly prefer \emph{after} at the start of the sentence, consistent with a preference for events to be mentioned chronological order. However, the models don't show the same level of preference for \emph{before} in the middle of the sentence, showing instead more of an even split between the two connectives. The ``Pair'' column indicates the percentage of pairs in which models show the target preference on both items. Half of models fall just about at chance level of 25\%. The other half of models exceed this chance-level percentage---particularly the BERT models and \AB{}. However, the percentages never exceed 41\%, suggesting that while some models may trend a bit toward this temporal ordering preference in their connective predictions, this trend is fairly weak. 

As we have established above, models' lack of the target trend in this experiment cannot be considered ``incorrect'' in any sense---however, it does indicate that models lack yet another humanlike pattern with respect to processing of discourse connectives, where such a pattern would have suggested finer-grained sensitivity to temporal dynamics. 



\section{Discussion}

In the above experiments, we have studied the competence of pre-trained LMs in predicting and interpreting connectives. Examining connective prediction in naturally-occurring data suggests that models may have certain go-to connectives that they predict across a variety of contexts. So we turn to more controlled tests inspired by psycholinguistics, to examine the extent to which models' handling of connectives reflects pragmatic competence. The results of these controlled tests suggest that models are not yet equipped to use subtle pragmatic cues to inform connective predictions---whether the test involves distinguishing causal from concessive discourse relations in settings of high syntactic and lexical overlap, or making predictions based on inference, reinforcement, or cancellation of implicatures. We also find that models appear to be insensitive to redundancy or contradiction, and that although certain models may have a slight tendency to prefer connectives that suggest chronological event mentions, this tendency is weak at best.

Variation between models is fairly minor, though some RoBERTa and ALBERT variants at times distinguish themselves in coming closer to chance-level performance when other models are close to 0\% accuracy. This could be attributable to larger pre-training data in the case of RoBERTa, and in the case of ALBERT, we speculate that some benefit may be derived from the sentence order prediction loss~\cite{Lan2020ALBERTAL}, which may encourage sensitivity to certain discourse dynamics. However, it is important to note that in these cases the models still at best barely surpass chance-level performance, so this does not indicate any particularly strong pragmatic competence from these models.

Why do models perform so poorly on the controlled tests? It is clear, of course, that these models have not been trained to do these specific tasks. However, if models have competence in pragmatic reasoning, that competence should be reflected in the preferences and distinctions tested for here. Our use of minimal pairs creates a particular challenge, in reducing models' capacity to succeed on the basis of shallower types of cues---syntactic, lexical---that they are more likely to have learned to prioritize. However, this means that these tests hopefully give us a clearer look at models' ability to use pragmatic cues per se. All in all, our results point to a situation in which sophisticated pragmatic reasoning is not yet a property of current models---at least not the aspects of pragmatics tested for here.

What are the implications of these results for our approach to these models? In terms of gauging linguistic competence that emerges from pre-training, the results suggest that word prediction objectives alone may not suffice to force models to learn nuances of pragmatic reasoning. Models tested here, at least, do not suggest that such pragmatic reasoning has been learned. The results on naturally-occurring data may suggest one source of pre-training limitations: models may be able to achieve reasonable prediction outcomes simply by defaulting to versatile connectives in a wide range of contexts, and by relying on lower-level syntactic and lexical cues. For models to learn pragmatics, it may be necessary to reduce reliability of shallower cues, scaffold learning with more meaning-rich supervision, or both. Enriched pragmatic competence may be achievable through fine-tuning of pre-trained models, but a fine-tuning approach will be subject to the same risks of models defaulting to shallower heuristics---so the same considerations will apply. We leave the problem of improving models' pragmatic competence for future work.   

\section{Conclusion}
The above experiments have examined pragmatic competence in pre-trained language models, with respect to discourse connectives. Results suggest that models are not yet equipped to use high-level pragmatic cues or reasoning to guide predictive behaviors, even if they show reasonable predictive accuracy in naturally-occurring data. We suggest that arriving at more pragmatically competent models may require greater control of shallow cues, or use of more meaning-rich training signal. We hope that this work will help to shed light on the linguistic competence of pre-trained LMs, and ultimately contribute to advancement in the pragmatic competence of models in NLP.


\section*{Acknowledgments}

We would like to thank three anonymous reviewers for helpful comments and suggestions. This material is based upon work supported by the National Science Foundation under Award No.~1941160.

\bibliography{anthology,custom}
\bibliographystyle{acl_natbib}


\section{Appendix}\label{sec:appendix}
\begin{table*}[t!]
\centering
\begin{tabular}{l|l l l l}
\toprule
  Connective & \BB & \BL & \RB & \RL \\
  \midrule
after & 0.0206 & 0.0239 & 0.0546 & 0.0475 \\
also & 0.0278 & 0.0282 & 0.0284 & 0.0305 \\
although & 0.0126 & 0.0147 & 0.0068 & 0.0097 \\
and & \textbf{0.4834} & \textbf{0.4591} & \textbf{0.3503} & \textbf{0.34} \\
as & 0.0623 & 0.0616 & 0.0819 & 0.0928 \\
because & 0.068 & 0.0722 & 0.074 & 0.0659 \\
before & 0.0098 & 0.0102 & 0.0122 & 0.0153 \\
but & 0.0354 & 0.0461 & 0.0927 & \textbf{0.1106} \\
for & 0.0172 & 0.0149 & 0.0098 & 0.0067 \\
if & 0.0567 & 0.0427 & 0.0373 & 0.0265 \\
since & 0.0091 & 0.0108 & 0.0071 & 0.0104 \\
so & 0.009 & 0.0113 & 0.0149 & 0.0114 \\
still & 0.0121 & 0.011 & 0.0133 & 0.0111 \\
then & 0.0108 & 0.0123 & 0.0109 & 0.0107 \\
though & 0.0037 & 0.0031 & 0.0128 & 0.0151 \\
until & 0.01 & 0.0115 & 0.008 & 0.0084 \\
when & 0.0844 & 0.0769 & 0.0625 & 0.0591 \\
while & 0.037 & 0.0521 & 0.0727 & 0.0825 \\
yet & 0.0012 & 0.0019 & 0.0028 & 0.0027 \\

  \bottomrule
\end{tabular}
\caption{Error percentage of connectives for BERT and RoBERTa family}
\label{table:error_percentage_of_connectives_PDTB_bert}
\end{table*}
\begin{table*}[t!]
\centering
\begin{tabular}{l|l l l l}
\toprule
  Connective &\AB &\AL&\AXL&\AXXL \\
  \midrule
after & 0.0457 & 0.0782 & 0.0511 & 0.0263 \\
also & 0.0281 & 0.0275 & 0.0361 & 0.0161 \\
although & 0.0096 & 0.0215 & 0.0437 & \textbf{0.1} \\
and & 0.0285 & 0.067 & \textbf{0.15} & 0.0512 \\
as & 0.018 & 0.0373 & 0.0409 & 0.0102 \\
because & \textbf{0.11} & \textbf{0.23} & \textbf{0.19} & \textbf{0.15} \\
but & \textbf{0.16} & \textbf{0.16} & \textbf{0.12} & \textbf{0.09} \\
for & 0.0135 & 0.0094 & 0.0153 & 0.0049 \\
if & 0.0565 & 0.0504 & 0.047 & 0.0232 \\
separately & 0.0041 & 0.0012 & 0.0023 & 0.0127 \\
since & 0.0069 & 0.0374 & 0.0127 & 0.0098 \\
so & 0.0022 & 0.0067 & 0.0097 & 0.0058 \\
specifically & 0.0001 & 0.0011 & 0.0003 & 0.0013 \\
still & 0.0166 & 0.0141 & 0.0119 & 0.0124 \\
then & 0.0156 & 0.0113 & 0.0118 & 0.0069 \\
though & 0.0053 & 0.0085 & 0.0071 & 0.0062 \\
ultimately & 0.0002 & 0.0014 & 0.0009 & 0.0048 \\
unless & \textbf{0.17} & 0.0456 & 0.0121 & 0.0619 \\
until & 0.0057 & 0.0136 & 0.0095 & 0.0128 \\
when & \textbf{0.10} & 0.0626 & \textbf{0.09} & \textbf{0.09} \\
whereas & 0.0228 & 0.0116 & 0.0023 & 0.0343 \\
while & \textbf{0.11} & 0.0469 & 0.0622 & \textbf{0.12} \\

\bottomrule
\end{tabular}
\caption{Error percentage of connectives for ALBERT family}
\label{table:error_percentage_of_connectives_PDTB_albert}
\end{table*}

\begin{table*}[ht]
\centering
\begin{tabular}{p{.2\textwidth}|p{.7\textwidth}}
\toprule
Condition & Example item \\
\midrule
Reinforcement test & \emph{The wind dispersed the sheep \underline{\textbf{and}} the wolves seized a lamb, which is to say, the wind dispersed the sheep [MASK] the wolves seized a lamb.} \textbf{(ideal target probs: \emph{before} $\gg$ \emph{after})}\\  
 & \emph{The wind dispersed the sheep \underline{\textbf{and then}} the wolves seized a lamb, which is to say, the wind dispersed the sheep \underline{\textbf{[MASK]}} the wolves seized a lamb.} \textbf{(ideal target probs: \emph{before} $>$ \emph{after})}\\ \midrule
Cancellation test & \emph{The wind dispersed the sheep \underline{\textbf{and}} the wolves seized a lamb, in fact the wind dispersed the sheep \underline{\textbf{[MASK]}} the wolves seized a lamb.} \textbf{(ideal target probs: \emph{after} $\gg$ \emph{before})}\\ 
 & \emph{The wind dispersed the sheep \underline{\textbf{and then}} the wolves seized a lamb, in fact the wind dispersed the sheep \underline{\textbf{[MASK]}} the wolves seized a lamb.} \textbf{(ideal target probs: \emph{after} $>$ \emph{before})}\\
\bottomrule
\end{tabular}
\caption{Example items for testing models' sensitivity to redundancy and contradiction, using implicature reinforcement and cancellation environments ($\gg$ (models assigned probability) much bigger than; $>$ bigger than)}\label{tab:reincancelitemssup}
\end{table*}
\begin{table*}[t!]
\centering
\begin{tabular}{l|l|l|l}
\toprule
  Model & Cancellation & Reinforcement & Pair\\
  \midrule
  \BB & 0.0063 & 0.0125&0.0093\\
  \BL & 0&0&0\\
  \RB & 0&0&0\\
  \RL & 0.0063&0.21&0.0031\\
  \AB & 0.025& 0.006& 0.02\\
  \AL & 0& 0&0\\
  \AXL & 0.01& 0.03& 0.02\\
  \AXXL &0 & 0.07& 0.04\\
  \bottomrule
\end{tabular}
\caption{Accuracy in tests of sensitivity to contradiction and redundancy in cancellation and reinforcement environments}
\label{table:Acc_Modified_Drenhaus_Cancellability}
\end{table*}

\end{document}